\documentclass[10pt,twocolumn,letterpaper]{article}
\usepackage{cvpr}              
\pdfoutput=1
\makeatletter
\@namedef{ver@everyshi.sty}{}
\makeatother
\usepackage{graphicx}
\usepackage{amsmath}
\usepackage{amssymb}
\usepackage{booktabs}
\RequirePackage{xcolor}
\usepackage[misc]{ifsym}
\usepackage{siunitx}
\usepackage{gensymb}
\usepackage{tcolorbox} 
\usepackage{stfloats} 
\usepackage{multirow}
\usepackage{dcolumn}
\usepackage{tcolorbox} 
\usepackage{stfloats} 
\usepackage{caption} 
\usepackage{float}
\usepackage{amsmath,amssymb,amsfonts}
\usepackage{algorithmic}
\usepackage{textcomp}
\usepackage{xcolor}
\usepackage{notoccite}
\usepackage{colortbl}
\usepackage{comment,mfirstuc}
\usepackage{cite}

\newcommand{\addComment}[2]{
  \expandafter\newcommand\csname #1\endcsname[1]{{\bf \color{#2} \capitalisewords{#1}:\,##1}}
  \expandafter\newcommand\csname #1cor\endcsname[2]{{\color{#2} \capitalisewords{#1}:\,\st{##1}{\bf ##2}}}
  \expandafter\newcommand\csname #1color\endcsname{#2}
}
\addComment{james}{blue} 
\addComment{martin}{red}
\addComment{robert}{green}
\usepackage[pagebackref=true,breaklinks=true,colorlinks,bookmarks=false]{hyperref}
\usepackage[capitalize]{cleveref}
\crefname{section}{Sec.}{Secs.}
\Crefname{section}{Section}{Sections}
\Crefname{table}{Table}{Tables}
\crefname{table}{Tab.}{Tabs.}

\begin{document}

\title{T-FFTRadNet:  Object Detection with Swin Vision Transformers from Raw
ADC Radar Signals}
\author{James Giroux$^1$, Martin Bouchard$^1$, Robert Laganiere$^{1,2}$ \\
University of Ottawa$^1$\\
Sensor Cortek Inc.$^2$ \\
Ottawa, Ontario, Canada\\
{\tt\small \{jgiro037, bouchm, laganier\}@uottawa.ca}
}

\maketitle

\begin{abstract}
Object detection utilizing Frequency Modulated Continous Wave radar is becoming increasingly popular in the field of autonomous systems. Radar does not possess the same drawbacks seen by other emission-based sensors such as LiDAR, primarily the degradation or loss of return signals due to weather conditions such as rain or snow. However, radar does possess traits that make it unsuitable for standard emmision-based deep learning representations such as point clouds. Radar point clouds tend to be sparse and therefore information extraction is not efficient. To overcome this, more traditional digital signal processing pipelines were adapted to form inputs residing directly in the frequency domain via Fast Fourier Transforms. Commonly, three transformations were used to form Range-Azimuth-Doppler cubes in which deep learning algorithms could perform object detection. This too has drawbacks, namely the pre-processing costs associated with performing multiple Fourier Transforms and normalization. We explore the possibility of operating on raw radar inputs from analog to digital converters via the utilization of complex transformation layers. Moreover, we introduce hierarchical Swin Vision transformers to the field of radar object detection and show their capability to operate on inputs varying in pre-processing, along with different radar configurations, \textit{i.e.} relatively low and high numbers of transmitters and receivers, while obtaining on par or better results than the state-of-the-art.

\end{abstract}

\section{Introduction}

With the capability to operate in adverse weather conditions, the prevalence of Frequency Modulated Continuous Wave (FMCW) radar within the field of autonomous driving is steadily growing. Moreover, various deep-learning architectures have been proposed to localize and classify objects within pre-processed return signals, a step beyond traditional radar applications \cite{Rebut_2021,Zhang_2021,Cao_2021,Decourt_2022}. Pre-processing generally consists of multiple Fast Fourier Transforms (FFT), producing combinations of range, azimuth, and Doppler information in what is known as RAD cubes. Processing of the full RAD cube is computationally expensive, and undesirable if the intended deployment is on low-power hardware. It is possible to utilize `Point-Cloud' (PC) representations of radar signals, in which the peaks in the range-azimuth spectrum can be converted to Cartesian reflection points. While utilizing LiDAR point clouds within deep learning frameworks has been successful, the task becomes much more difficult with radar given the sparsity of points.
As such, a more desirable approach is to minimize the number of FFTs used, in which range-Doppler spectra are suitable candidates. Ideally, one is able to utilize raw analog to digital converter (ADC) inputs and remove pre-processing entirely. Utilizing the raw ADC values also allows the network to potentially learn the characteristics of the radar sensor, which could be crucial given low resolution and high noise configurations. In this work we present T-FFTRadNet, building off the prior work of Rebut et al. \cite{Rebut_2021}, in which we show the robustness of Vision Transformers \cite{Liu_2021} as feature extractors in object detection networks. Specifically, the capability of Swin Vision Transformers to operate on High Definition (HD) radar, Low Definition (LD) radar and varying degrees of pre-processing, \textit{i.e.} raw ADC, range-Doppler matrices and full RAD cubes. We utilize the learnable transformations from Zhao et al. \cite{Zhao_2023} and show the benefits in low resolution radar applications. Experiments are performed on the LD RADDet dataset \cite{Zhang_2021} and the HD RadIal dataset \cite{Rebut_2021}. 

\section{Related Works}

Radar object detection generally requires some degree of pre-processing, usually in the form of FFTs. The utilization of multiple FFTs across Range-Doppler-Azimuth axes allows efficient information extraction in the frequency domain. Transformations across subsequent axes also cause network complexity to scale, \textit{i.e.} when the full RAD cube is utilized. Thus, the ideal radar network is one which reduces the number of frequency transformations required. Rebut et al.\cite{Rebut_2021} show that networks utilizing RD tensors (FFTRadNet) can approximate this third transformation and efficiently extract angular information. Moreover, they are able to further reduce complexity via the usage of coarse RA grids and regression heads. The center of objects are predicted within a grid $\frac{1}{8}$ and $\frac{1}{4}$ the native azimuth and range resolutions. A secondary regression head then predicts fine-grained correction factors producing localization well within the physical sensor resolutions. Yi et al. \cite{Yi_2023} further build on this idea, utilizing a Cross-Modal supervision strategy. Manual annotation of radar signals is difficult due to their low interoperability and high noise levels. The authors in \cite{Yi_2023} develop a novel FCN style network operating on range-Doppler inputs, which is trained under the supervision of radar-adapted, image segmentation networks. This method alleviates manual annotation and is immune to false targets due to noise. We emphasize that this method uses `improved' labels when comparisons are presented in later sections. A further study could be the implementation of this training strategy with our Swin based network.

Object detection directly on RD maps has proven to be fruitful, in which the task generally becomes one of segmentation \cite{Decourt_2022,Ouaknine_2021_ICCV,Shirakata_2019,Ouaknine_2020}. U-Net \cite{ronneberger_2015} and FCN \cite{Long_2015} like structures are suitable candidates for such implementation backbones. While this does not provide direct angular information, it allows the extraction of Regions of Interest (ROIs) in the discrete range and Doppler bins. It is then possible to perform the azimuthal FFT over only subsets of the entire signal and extract their angular information. Generally, the input RD maps are integrated over the antenna axis to reduce network complexity. It is also possible to combine this information with RA maps \cite{Ouaknine_2021_ICCV}, known as \textit{Multi-view Methods}, which are integrated over the chirp, or slow time axis.

RAD cubes contain range-Doppler and azimuth signatures of objects, and are therefore the most information-dense, allowing efficient object detection \cite{Zhang_2021,Palffy_2020,Xiangyu_2021}. They are also computationally expensive to produce and the resulting networks are more complex given the higher dimensional inputs.

The most common practice in recent years is to utilize PC representations \cite{Svenningsson_2021,Nabati_2021,Wang_2020,Prophet_2020,Danzer_2019,Meyer_2019}. This is due to the successes in LiDAR PC object detection. Radar PC representations suffer from pre-processing costs along with loss of information due to filtering. This creates sparse PC representations not always suitable for object detection tasks.

\section{Proposed Approach}

\subsection{Hierarchical Vision Transformers as Feature Extractors}

\begin{figure*}
    \centering
    \includegraphics[width=\textwidth]{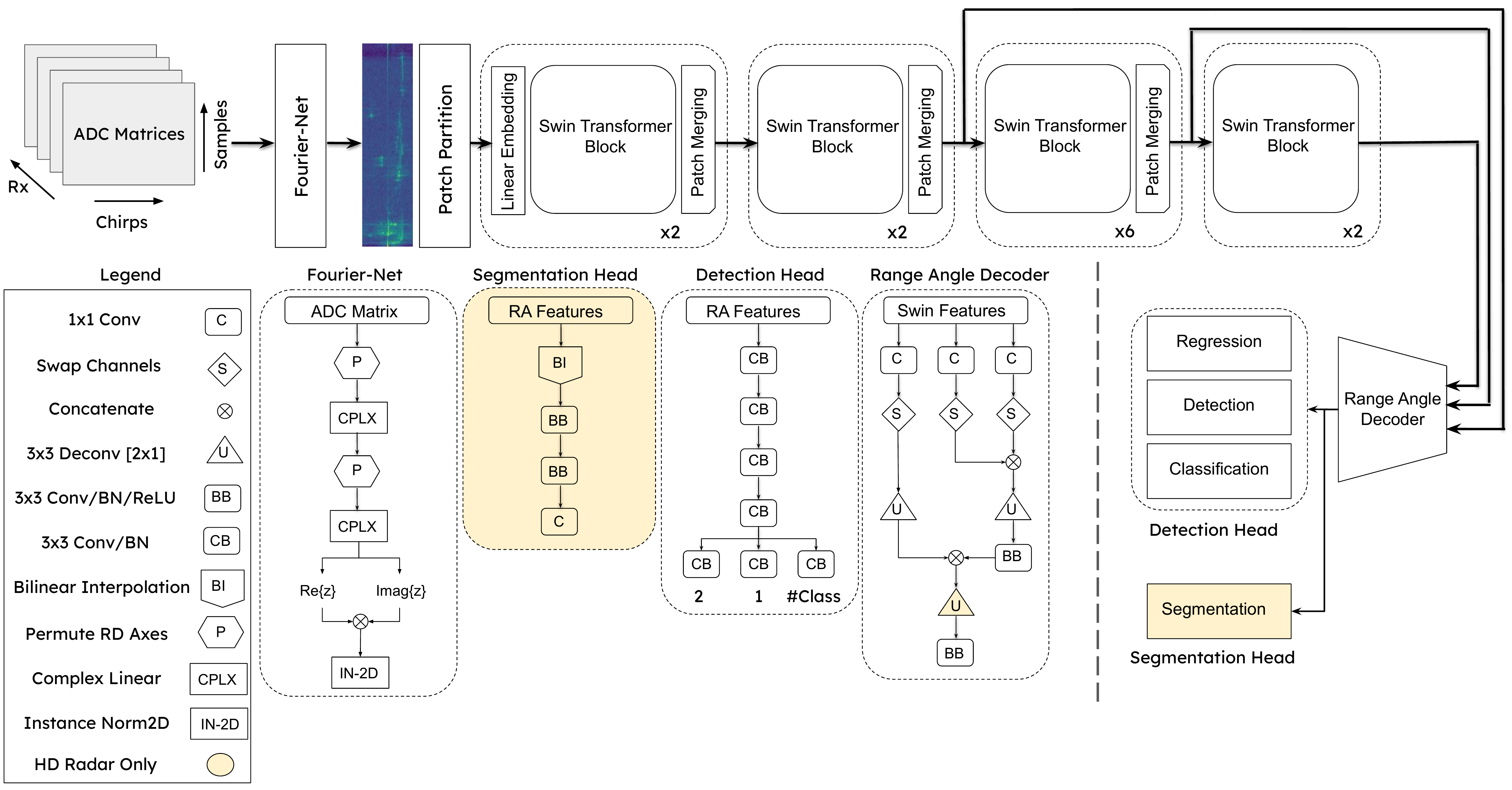}
    \caption{\textbf{Network Structure:} Overall network structure utilizing raw ADC as input. RD inputs replace the Fourier-Net with range-Doppler FFTs. RAD cube inputs replace the Fourier-Net with range-Doppler-azimuth FFTs and remove channel swapping operations in the Range-Angle Decoder. }
    \label{fig:model_diagram}
\end{figure*}

We build off the architecture proposed in Rebut et al. \cite{Rebut_2021}, in which our main contributions are the application of vision transformers to the field of radar object detection, along with the adaption of raw ADC as input. We replace the front end of the network with a hierarchical Swin Transformer \cite{Liu_2021}. It is well known that the performance of vision transformers scales with available data. Thus, we deem this research timely as increasingly large radar datasets become available. The overall network structure is shown in Fig.\ref{fig:model_diagram}.

Utilizing a hierarchical vision transformer we are able to extract more prominent features, of varying resolution, to be fed to the network back-end. This is accomplished via the utilization of $2\times 2$ patches, significantly smaller than $3\times 3$ convolutional windows commonly seen in other Feature Pyramid Networks (FPN), or FCN style networks operating on radar. Moreover, the hierarchical nature of a Swin transformer allows the aggregation of varying resolution features via patch merging between consecutive transformer blocks. In downstream Swin blocks, Windowed Multi-Head Self-Attention (W-MHSA) allows the network to retain linear complexity with respect to the input. This is due to the fact that self-attention is only computed in local neighborhoods (windows), as opposed to the quadratic complexity required by standard vision transformers when self-attention is computed across all available patches. Sequentially connected layers shift said windows, providing connections between them \cite{Liu_2021}. Eq.\ref{eq:SW_MHSA} depicts the flow of outputs between two consecutive blocks, $x^k$ and $x^{k+1}$, where $\hat{x}$ and $x$ denote the outputs of the W-MHSA, Shifted Window Multi-Head Self-Attention (SW-MHSA), LayerNorm (LN) and Multi-layer Perceptron (MLP) modules.

\begin{equation}
    \label{eq:SW_MHSA}
    \begin{aligned}
        \hat{\mathbf{x}}^l = & \, \text{W-MHSA}(\text{LN}(\mathbf{x}^{l-1})) + \mathbf{x}^{l-1} \\
        \mathbf{x}^l = & \, \text{MLP}(\text{LN}(\hat{\mathbf{x}}^l)) + \hat{\mathbf{x}}^l \\
        \hat{\mathbf{x}}^{l+1} = & \, \text{SW-MHSA}(\text{LN}(\mathbf{x}^{l})) + \mathbf{x}^{l} \\
        \mathbf{x}^{l+1} = & \, \text{MLP}(\text{LN}(\hat{\mathbf{x}}^{l+1})) + \hat{\mathbf{x}}^{l+1} 
    \end{aligned}
\end{equation}

The model is designed to take primarily RD or raw ADC matrices as input, although we show the utilization of RAD cubes in select cases. When raw ADC inputs are utilized, we include transformation layers prior to the Swin transformer. These are discussed in more detail in Sec.\ref{subsec:ADC}. Being that transformer style networks utilize functional embeddings, the utilization of varying inputs (such as RD, RAD, etc.) does not drastically increase network complexity. 

The Swin transformer extracts varying resolution features which are then sent to the back end of the pre-existing FFTRadNet network. First, the Range-Angle Decoder transforms these features from a compressed RD representation to RA. These decoded features then feed the output heads. The detection head produces a binary grid in the RA coordinate system of the radar sensor such that occupied cells represent the center points of objects. In both HD and LD radar, these grids are reduced representations of the physical sensor limitations to alleviate computational complexity. To compensate for the reduction, the regression head predicts correction factors in both range and angle to allow finer grain resolution for detected objects. We train the binary detection and regression heads utilizing a weighted combination of Focal (detection) and \textit{smooth L1} loss (regression), Eq.\ref{eq:binary_head}.

\begin{equation}\label{eq:binary_head}
\begin{aligned}
    \mathcal{L}_{Bin}(\mathbf{x},\mathbf{y}_{b},\mathbf{y}_{r}) =  \alpha \text{Focal}(\mathbf{y}_{b},\hat{\mathbf{y}}_{b}) +  \\
    \beta \text{smooth-L1}(\mathbf{y}_{r} - \hat{\mathbf{y
}}_{r})
\end{aligned}
\end{equation}

 Where $\mathbf{y}_b \in [0,1]^{B_{R /n} \times B_{A/m}}$, $\mathbf{\hat{y}}_b \in {0,1}^{B_{R /n} \times B_{A/m}}$ and $\mathbf{y}_r \in \mathcal{R}^{B_{R /m} \times B_{A/n}}$, in which $m,n$ denote the range and angular reduction factors. For HD and LD radar, $m=4, n = 8$ and $m=4, n = 2$, respectively. $\alpha$ and $\beta$ are scaling factors set to $2$ and $10^2$.
 
In the case of LD radar we add an additional classification head, which produces a likelihood map containing $C$ channels, where $C$ is the number of classes.\footnote{In the case of the HD radar dataset, we do not have class labels.} Given a detection in the binary head, we can then obtain the corresponding classes of detected objects by taking the maximum argument of the likelihood maps at the localized regions subsequent to a \textit{Softmax} function. We train this head independently utilizing pixel-wise cross-entropy loss, Eq.\ref{eq:cross_entropy}. Where $p_k^j$ is the probability predicted by the model of a given class $k$, at a specific pixel $j$. Being that the majority of values in the likelihood masks are zeros the weighting parameter $\gamma$ must be set to a large value of $10^5$ to produce loss contributions on the same orders of magnitude during training.

\begin{equation}\label{eq:cross_entropy}
    \begin{aligned}
            \mathcal{L}_{Class}(\mathbf{x},\mathbf{y}_c) = & -\gamma \sum_j^J\sum_k^K y_k^j \log (p_k^j)
    \end{aligned}
\end{equation}

For the HD radar dataset, we also utilize the segmentation head to predict the free driving space available to the vehicle in a polar coordinate system. As such, the task is defined as a pixel-wise binary classification problem in which binary cross-entropy, Eq.\ref{eq:bce}, lends itself naturally. The scale factor $\lambda$ is set to $10^2$.

\begin{equation}\label{eq:bce}
    \mathcal{L}_{Free}(\mathbf{y}_{seg},\hat{\mathbf{y}}_{seg}) = \lambda \sum_i^{\frac{B_{R}}{2}}\sum_j^{\frac{B_{A}}{4}} \text{BCE} \left(\mathbf{y}(i,j),\hat{\mathbf{y}}(i,j) \right)
\end{equation}

The total loss contribution then becomes a linear combination of $\mathcal{L}_{Bin}$ and $\mathcal{L}_{Class}$ or $\mathcal{L}_{Free}$.

\subsubsection{ADC Inputs - Linear Transformations}\label{subsec:ADC}

Conventional signal processing on radar relies heavily on the Discrete Fouter Transform (DFT) algorithm, transforming high dimensional FMCW inputs from the time to frequency domain. This transformation also dominates in neural network (NN) based, radar object detection as localizing objects in the time domain is increasingly difficult. DFTs allow a network, specifically a vision-based network, to exploit the dominating frequency components (seen as bright spots) caused by object reflections in downstream tasks. However, utilizing conventional DFTs possess some inherent computational cost, along with the general requirement that some normalization should be performed prior to network injection. An ideal radar object detection network is then one that is capable of utilizing non-normalized inputs in the time domain and learning a transformation suitable to the task. Zhao et al. \cite{Zhao_2023} propose complex-valued linear layers, utilizing the prior knowledge of the Fourier transform. The layers can be seen as an $M\times M$ matrix, where each column acts as a learnable transformation over a slice of the input $\mathbf{x}$ of size $N\leq M$. Thus, mimicking the action of a standard DFT while allowing the network to identify potentially more optimal transformations. Note that permutations are required between consecutive layers to correctly orient the multiplication to mimic a DFT over range or Doppler axis. The weights of the layer are initialized as shown in Eq.\ref{eq:Fourier_weights}, with values corresponding to a standard DFT.

\begin{equation}\label{eq:Fourier_weights}
    w(k,m) = \exp{(-j\frac{2\pi}{N}km)}, \, 
    \begin{cases}
        0 \leq k \leq M-1 \\
        0 \leq m \leq M-1 
    \end{cases}
\end{equation}

It is also preferred that the transformation preserves the continuity of reflected objects in the learned Doppler space, \textit{i.e.} zero Doppler velocity should be centralized. This is done by shifting the upper and lower half of the weights and is equivalent to the commonly used \textit{fftshift} functions. Initializing the network in such a way provides an initial starting point that has been shown to work well for radar object detection, yet the transformation learned through end-to-end training is likely to no longer be Fourier.\cite{Zhao_2023}
In our network, we first inject non-normalized ADC signals to a range transformation layer, followed by a Doppler transformation layer. Given that DFTs in general can produce large values, we utilize a normalization layer before injection into the Swin Transformer. We also experiment with utilizing non-linear activation functions on the outputs of the range and Doppler layers, namely those inspired by Phase-Amplitude (PA) functions which preserve the phase of the activation value. A modified version of \textit{modReLU} \cite{Ariovsky_2016} is used, in which we utilize a \textit{LeakyReLU} to preserve negative values of the FFT yet provide non-linearity, Eq.\ref{eq:leakymodReLU}, where $z$ and $b$ are the input vector and bias, respectively.

\begin{equation}\label{eq:leakymodReLU}
    f(z) = \text{LReLU}(|z| + b)\frac{z}{|z|}
\end{equation}

In what follows, any experiment utilizing non-linear activation functions on ADC data will be denoted with an NL (Non-Linear) within the respective network title.

\section{Experiments}

\subsection{Experimental Setting}

\textbf{Datasets.} We utilize a single HD radar dataset, RadIal \cite{Rebut_2021} in which we perform generalized object detection. No class information is produced via the model, only a binary grid (1 being an object, 0 being no object) in a RA coordinate system. The HD radar sensor is composed of 12 transmitters ($T_x$) and 16 receivers ($R_x$), in which each $T_x$ produces 256 chirps, each sampled with 512 points. The inputs to the model are either ADC, or RD matrices in which the complex components of the I/Q data are appended as additional channels for the Swin Transformer, producing inputs of shape (512,256,$2R_x$). This utilization of complex components as additional channels to the Swin Transformer is consistent in all our experiments, except for RAD cubes where the norm of the cubes' values is used, resulting in all real-valued inputs.

For LD radar, we utilize RADDet \cite{Zhang_2021}. The sensor is composed of $2 \, T_x$ and $4 \, R_x$, where the transmitters produce 64 chirps each sampled 256 times producing inputs of shape (256,64,$T_x \cdot R_x$) where a virtual array has been formed. Utilizing the RADDet dataset, we show performance on ADC, RD and full RAD cube inputs for completeness. In LD radar the model predicts objects and their corresponding classes. In HD and LD datasets that utilize RD or RAD inputs, we normalize across respective channels via Eq.\ref{eq:Normalization}, where $i,j$ denote the range-Doppler bin and and $\mu_k$, $\sigma_k$ are the mean and standard deviation of the $k^{th}$ channel. We again note that ADC inputs possess no prior normalization.

\begin{equation}
\label{eq:Normalization}
     N(i,j,k)  =  \frac{M(i,j,k) - \mu_k}{\sigma_k} \\
\end{equation}

We pre-process the raw ADC for the RADDet dataset, producing RD matrices and RAD cubes in the data-loading pipeline via the Intel\textregistered~MKL numpy interface. RD axes are subject to Hamming windowing functions as shown in Eq.\ref{eq:window_functions}, reducing side lobe profiles. Where $N$ and $M$ represent the number of samples per chirp and number of chirps, respectively. Experiments are performed with and without windows for the RADDet and RadIal datasets. Note that by default, RadIal has performed windowing.

In both RADDet and RadIal, we perform experiments with and without centering around the zero frequency bin in the Doppler axis (utilizing an FFT shift). When RAD cubes are utilized in RADDet, we also shift the azimuth axis and remove the swapping of channels in the decoder given that inputs are already aligned properly with the desired output axis. Furthermore, we must modify the convolutions slightly to account for the symmetrically shaped output of the Swin transformer. The training and testing split utilized is provided in Zhang et al. \cite{Zhang_2021}.

\begin{equation}
\label{eq:window_functions}
\begin{aligned}
    W_{R} & = 0.54 - 0.46 \cos{ \left(2\pi \frac{n}{n-1}\right)},\; \, 0 \leq n \leq N\\
    W_{D} & = 0.54 - 0.46 \cos{ \left( 2\pi \frac{m}{m-1} \right)},\; 0 \leq m \leq M
\end{aligned}
\end{equation}

\textbf{Architecture specifications and computing resources.}

We utilize PyTorch \textit{1.12.1} \cite{pytorch_2019} as the deep learning framework. All code is built upon the framework provided by Rebut et al.\cite{Rebut_2021} All training and inference are done utilizing a single Nvidia RTX 3090 24GB card and Intel~\textregistered ~i7 12700K CPU.
Models are trained for 150 epochs total utilizing the Adam optimizer with an initial learning rate of $10^{-4}$ and a decay of $0.9$ every 10 epochs. The optimal model is selected via an F1 score for the HD radar, and by maximizing performance over both heads (classification, general object detection) for LD radar. In both cases, this is done utilizing the validation dataset. For the HD radar, a batch size of 4 is utilized, where as the LD radar we utilize batches of size 16 on RD inputs. For the full, padded RAD cube, we utilize batches of size 4. Table~\ref{tab:flops} contains information regarding network complexity and computational requirements for a single inference. Values for our architecture and FFTRadNet are calculated utilizing the \textit{DeepSpeed} \cite{Rasley_2020} framework, Cross Modal DNN is taken from \cite{Yi_2023} and RADDet is calculated using TensorFlow's \cite{tensorflow2015-whitepaper} profiling functions.

\begin{table}
\centering
\scalebox{0.78}{
\begin{tabular}{ c c c c } 
\hline
\textbf{Model} & \textbf{Dataset} & \textbf{FLOPs} $\downarrow$  & \textbf{\# Params.} $\downarrow$ \\
\hline
FFTRadNet \cite{Rebut_2021}       & RadIal   & 288 G & 3.79 M \\
Cross Modal DNN \cite{Yi_2023} & RadIal & 358 G & 7.7 M \\
T-FFTRadNet & RadIal  & 194 G & 9.64 M  \\
T-FFTRadNet & RADDet & 10 G &  7.5 M \\
T-FFTRadNet (RAD) & RADDet & 26 G & 8.22 M \\
RADDet (RAD) \cite{Zhang_2021} & RADDet & 25G & 9.59 M \\
\hline
\end{tabular}}
\scalebox{0.7}{
\begin{tabular}{c c c c } 
\multicolumn{4}{c}{\textbf{Fourier Transforms} } \\
\hline
\textbf{Model} & \textbf{Dataset} & \textbf{FLOPs} $\downarrow$  & \textbf{\# Params.} $\downarrow$ \\
\hline
Fourier-Net & RadIal & 12.9 G  & 327.68 K \\
RD FFT  & RadIal & 214 M  & N/A \\
Fourier-Net & RADDet & 337.6 M & 69.63k \\
RD FFT  & RADDet & 11 M  & N/A \\
\hline
\end{tabular}}
 \caption{\textbf{Computational Overhead:} Floating Point Operations per second (FLOPs) and network parameters. FLOPs is equated as $\sim 2\cdot$~MACs (Multiply and Accumulate Operations). Note that models utilizing ADC inputs will be the summation of the Fourier-Net and T-FFTRadNet.}   
 \label{tab:flops}
\end{table}

\subsection{Object Detection on RadIal (HD)}

Models are evaluated on the test set given in Rebut et al.\cite{Rebut_2021}, utilizing Average Precision (AP) and Average Recall (AR) for object detection, and mean Intersection over Union (mIoU) for free driving space segmentation. Table~\ref{tab:HD_Final_Comp} shows a collection of results for comparison between other models on the same dataset. The table also contains various renditions of our implementation operating on different inputs, or different pre-processing, \textit{i.e.} windowing or no windowing. Results with the best performance are presented in bold.

\begin{table*}
\centering
\setlength{\tabcolsep}{7pt}
\begin{tabular}{c c c c c c c}
\toprule
      &  AP ($\uparrow$) & AR($\uparrow$) & F1($\uparrow$) & Range Error($\downarrow$) & Angle Error($\downarrow$) & mIoU($\uparrow$) \\
\midrule
FFTRadNet \cite{Rebut_2021} &  $ \mathbf{96.8\%} $  & $82.2 \% $    & $ 88.9 \% $   & $ \mathbf{0.12m}$    & $ \mathbf{ 0.10 \degree} $    & $73.98 \% $   \\
Cross Modal DNN \cite{Yi_2023} &  $ \mathbf{96.9 \%} $  & $83.5 \% $    & $ \mathbf{89.7 \% } $   & $ N/A $    & $ N/A $    & $ \mathbf{80.4 \%} $   \\
T-FFTRadNet                             &$ 89.6 \% $  & $ \mathbf{89.5 \%} $    & $ \mathbf{89.5 \% }$   & $0.15m$    & $ \ang{0.12} $  &  $ \mathbf{80.2 \%} $   \\
T-FFTRadNet   (Shift)                          &$ 91.3 \% $  & $ 88.3 \% $    & $ \mathbf{89.8 \% }$   & $0.15m$    & $ \ang{0.12} $  &  $ 79.5 \% $   \\
T-FFTRadNet (No Windowing)     &  $89.4 \% $ & $87.6 \%$ & $88.5 \% $  &  $0.16m$ & $ \ang{0.12}$ & $79.3 \%$ \\
T-FFTRadNet (ADC)                            &$ 88.2 \% $  & $ 86.7 \% $    & $ 87.4 \% $   & $0.16m$    & $ \ang{0.13} $  &  $ 79.6 \% $   \\
T-FFTRadNet (ADC - NL)                            &$ 88.1 \% $  & $ 86.1 \% $    & $ 87.1 \% $   & $0.16m$    & $ \ang{0.13} $  &  $ 78.8 \% $ \\
T-FFTRadNet (ADC - No Shift)   & $ 88.1\% $   &   $85.5\% $   &    $86.7\% $  & $0.16m$  & $\ang{0.13}$  & $78.8\% $ \\
\bottomrule
\end{tabular}
\caption{\textbf{HD Radar - Architecture comparison:} Transformer based FFTRadNet (T-FFTRadNet), FFTRadNet \cite{Rebut_2021} and Cross Modal DNN \cite{Yi_2023}.Labels in brackets denote differing input types, or the addition/removal of pre-processing steps, \textit{i.e.}, removal of Hamming Window or utilizing an \textit{fftshift}.}
\label{tab:HD_Final_Comp}
\end{table*}

From Table~\ref{tab:HD_Final_Comp}, utilization of the Swin Transformer, on the standard RadIal data, provides significant increases in AR in comparison to FFTRadNet \cite{Rebut_2021} and Cross Modal DNN \cite{Yi_2023} by $\sim 6-7\%$, although AP generally lacks by $\sim 7\%$. It is interesting to note the equivalence between the two metrics (AP and AR) with our architecture, resulting in an F1 score on par with the state-of-the-art (SOTA) and a more balanced model in general. This is also the case regarding free driving space evaluation in which the mIoU is significantly higher than FFTRadNet by $\sim 6\%$ and on par with Cross Modal DNN. We also note that applying a Hamming window is an efficient method of improving performance and is consistent with results in later sections. Centering the zero frequency bin in the Doppler spectrum (denoted with Shift in Table~\ref{tab:HD_Final_Comp}) improves performance marginally in some metrics. It is likely not a crucial pre-processing step given that the amount of leakage with high-resolution data will be fairly low. Utilizing a shift should be investigated on a case-by-case basis.

Interestingly, we do not see increased performance via the utilization of raw ADC for HD radar. It is possible that in cases where resolution is not an issue, the utilization of a standard FFT is sufficient. Further optimization may yield parameter combinations that increase this performance. Being that network inputs are not normalized by design, the expected range is no longer static, which influences performance. We also note that the utilization of Non-Linear (NL) activation provides no benefit in our study and in fact, slightly inhibits performance likely due to the suppression of negative values. In either case, the transformation learned is no longer Fourier. The transformation contains a perturbation from a standard FFT.

\subsection{Object Detection on RADDet (LD)}

Models are evaluated on the test set given in Zhang et al.\cite{Zhang_2021}, in which the metric of choice is mean Average Precision (mAP). Table~\ref{tab:Swin_RADDet_AP} contains a comparison of our implementation, utilizing various inputs and the published results of RADDet \cite{Zhang_2021} and DAROD \cite{Decourt_2022}.

From Table~\ref{tab:Swin_RADDet_AP}, utilizing the transformer-based network on full RAD cubes significantly outperforms RADDet by $\sim 4\% $, and marginally outperforms DAROD utilizing RD inputs. We see a significant increase in performance over RD inputs when raw ADC is used, in terms of class-wise mAP (Table~\ref{tab:Swin_RADDet_AP} top). The model is able to learn a more optimal transformation (a perturbed FFT), along with the sensor characteristics of the radar sensor. The result is a model more capable of classifying objects within the radar spectrum. For LD radar datasets with shorter sequences in range, azimuth and Doppler dimensions, the resulting frequency domain resolution, referring to the Fourier transform main lobe widths, will be more limited. Utilizing data-dependent NN models can perhaps learn the average SNR of the data across each frequency and therefore provide a benefit that can compensate for the reduced resolution, up to some point.

Generalized object detection results are presented in Table~\ref{tab:RADDet_ObjectDet_Generalized}, utilizing the binary RA grid devised in FFTRadNet. The most optimal model with respect to general detection does not correspond to the highest mAP (all things equal), although the disparity between them is on the order of a few percent (ADC outperforms RD class-wise, RD outperforms in general detection). It is possible that loss contributions and their respective weighting play the reason for this discrepancy. Although both models utilize the same weightings, we have introduced additional transformation parameters with respect to ADC input. These parameters may be influenced in such a way that they better learn the differences between different object signatures, at the expense of general object feature extraction. Being that pixel-wise cross-entropy is subject to sparsity (large amounts of zeros), the loss is scaled in such a way that it is on equal orders of magnitude throughout training and therefore the trade-off should be partially minimized. Optimized loss weighting schemes, or perhaps the utilization of a different loss function could improve overall performance.

Table~\ref{tab:RADDet_classwise_AP} contains class-wise AP values for the models. From this table, it is apparent that the transformer network suffers specifically on the motorcycle class. This is due to the fact that the class is underrepresented within the dataset causing these to be commonly classified as bicycles or cars depending on their relative position and angle with respect to the sensor.

\begin{table*}
\centering
\begin{tabular}{c c c c c c}
\multicolumn{6}{c}{\textbf{RD and ADC Inputs} } \\
\toprule
& T-FFTRadNet & T-FFTRadNet (NL) & T-FFTRadNet & T-FFTRadNet & Decourt et al.\cite{Decourt_2022} \\
\midrule
 Input & ADC & ADC & RD & RD (No Windowing) & RD \\
 mAP($\uparrow$) & $\mathbf{49.1\%}$ & $45.6\%$ & $46.8\%$ & $44.6\%$ & $46.6\%$ \\
 \bottomrule
\end{tabular}

\begin{tabular}{c c c}
\multicolumn{3}{c}{\textbf{RAD Inputs} } \\
\toprule
& T-FFTRadNet & Zhang et al.\cite{Zhang_2021}  \\
\midrule
 Input & RAD & RAD  \\
 mAP($\uparrow$) & $\mathbf{55.7\%}$ & $51.6\%$ \\
 \bottomrule
\end{tabular}
\caption{\textbf{LD Radar - RADDet Results:} mean Average Precision over the six object classes. Sub-tables  correspond to models in which complexity and input structure are the same, ADC and RD (top) and RAD cube (bottom).}
\label{tab:Swin_RADDet_AP}
\end{table*}

\begin{table}
\centering
\scalebox{0.82}{
\begin{tabular}{c c c c c c c }

\multicolumn{7}{c}{\textbf{T-FFTRadNet (ADC - NL)} } \\
\toprule
          & bicycle &   bus    & car & motorcycle & person  &   truck \\
\midrule
AP     &    $56.6\%$            &  $37.5 \%$       &     $62.6 \% $              &      $0.0 \%  $         &  $66.5 \% $           & $50.5 \% $         \\
AR     &    $ 29.4 \% $         & $ 7.9 \% $       &     $ 43.6 \% $              &      $ 0.0 \% $         & $ 36.9 \% $           & $ 30.4 \% $   \\
\bottomrule

\multicolumn{7}{c}{\textbf{T-FFTRadNet (ADC)} } \\
\toprule
          & bicycle &   bus    & car & motorcycle & person  &   truck \\
\midrule
AP     &    $62.5\%$            &  $50.0 \%$       &     $63.0 \% $              &      $0.0 \%  $         &  $70.7 \% $           & $48.4 \% $         \\
AR     &    $ 31.9 \% $         & $ 10.5 \% $       &     $ 44.5 \% $              &      $ 0.0 \% $         & $ 38.8 \% $           & $ 31.7 \% $   \\
\bottomrule

\multicolumn{7}{c}{\textbf{T-FFTRadNet (RD)} } \\
\toprule
          & bicycle &   bus    & car & motorcycle & person  &   truck \\
\midrule
AP     &    $72.6\%$            &  $20.0 \%$       &     $65.1 \% $              &      $0.0 \%  $         &  $71.8 \% $           & $51.4 \% $         \\
AR     &    $ 29.9 \% $         & $ 5.3 \% $       &     $ 50.6 \% $              &      $ 0.0 \% $         & $ 39.6 \% $           & $ 35.4 \% $   \\
\bottomrule

\multicolumn{7}{c}{\textbf{T-FFTRadNet (RD - No Windowing)} } \\
\toprule
          & bicycle &   bus    & car & motorcycle & person  &   truck \\
\midrule
AP     &    $62.3\%$            &  $28.6 \%$       &     $60.6 \% $              &      $0.0 \%  $         &  $70.6 \% $           & $45.7 \% $         \\
AR     &    $ 21.1 \% $         & $ 5.3 \% $       &     $ 37.5 \% $              &      $ 0.0 \% $         & $ 33.5 \% $           & $ 27.1 \% $   \\
\bottomrule


\multicolumn{7}{c}{\textbf{T-FFTRadNet (RAD)} } \\
\toprule
          & bicycle &   bus    & car & motorcycle & person  &   truck \\
\midrule
AP     &    $60.1\%$            &  $50.0 \%$       &     $64.6 \% $              &      $33.3 \%  $         &  $70.6 \% $           & $55.9 \% $         \\
AR     &    $ 34.8 \% $         & $ 18.4 \% $       &     $ 50.5 \% $              &      $ 4.8 \% $         & $ 44.9 \% $           & $ 39.9 \% $   \\
\bottomrule
\end{tabular}}
\caption{\textbf{LD Radar - RADDet Classwise AP and AR:} Notice the decrease in performance across specific classes. The cause of this is twofold, the first reason being that certain classes are underrepresented in the dataset, and the second is due to similar signatures in the RD spectrum from certain objects.}
\label{tab:RADDet_classwise_AP}
\end{table}

\begin{table}
\centering
\scalebox{.90}{
\begin{tabular}{c c c c }
\toprule
Architecture & Input & AP ($\uparrow$) & AR ($\uparrow$) \\
\midrule
T-FFTRadNet           &       ADC (NL)           & $59.9 \% $    &    $ 54.1 \%$ \\
T-FFTRadNet           &       ADC            & $  61.8\% $    &   $ 54.5 \% $   \\ 
T-FFTRadNet           &       RD            & $\mathbf{  64.5\%} $    &   $\mathbf{ 57.5 \%} $   \\   
T-FFTRadNet           &    RD (No Windowing) & $ 58.5 \% $   &    $ 48.4 \% $ \\
T-FFTRadNet           &       RAD         & $  \mathbf{64.2\%} $    &   $ \mathbf{56.5 \%} $   \\    
\bottomrule
\end{tabular}}
\caption{\textbf{LD Radar - RADDet Generalized Detection Head Results:} Results from the generalized object detection head, utilizing IoU and NMS thresholds of 0.5 and 0.3, respectively.}
\label{tab:RADDet_ObjectDet_Generalized}
\end{table}

\section{Discussion}

We have shown that Vision Transformers, namely hierarchical Swin Transformers are capable of forming the backbone for efficient object detection on varying radar inputs. Given that transformers utilize primarily dense connections and embeddings, the computational throughput required for a single inference pass is significantly lower (see FLOPs in Table~\ref{tab:flops}) than convolutional based feature extractors. Moreover, the adaptation to differing inputs (RD vs RAD for example) requires fewer network alterations given fixed-sized embeddings.

\begin{figure}
    \centering
    \includegraphics[height=11cm]{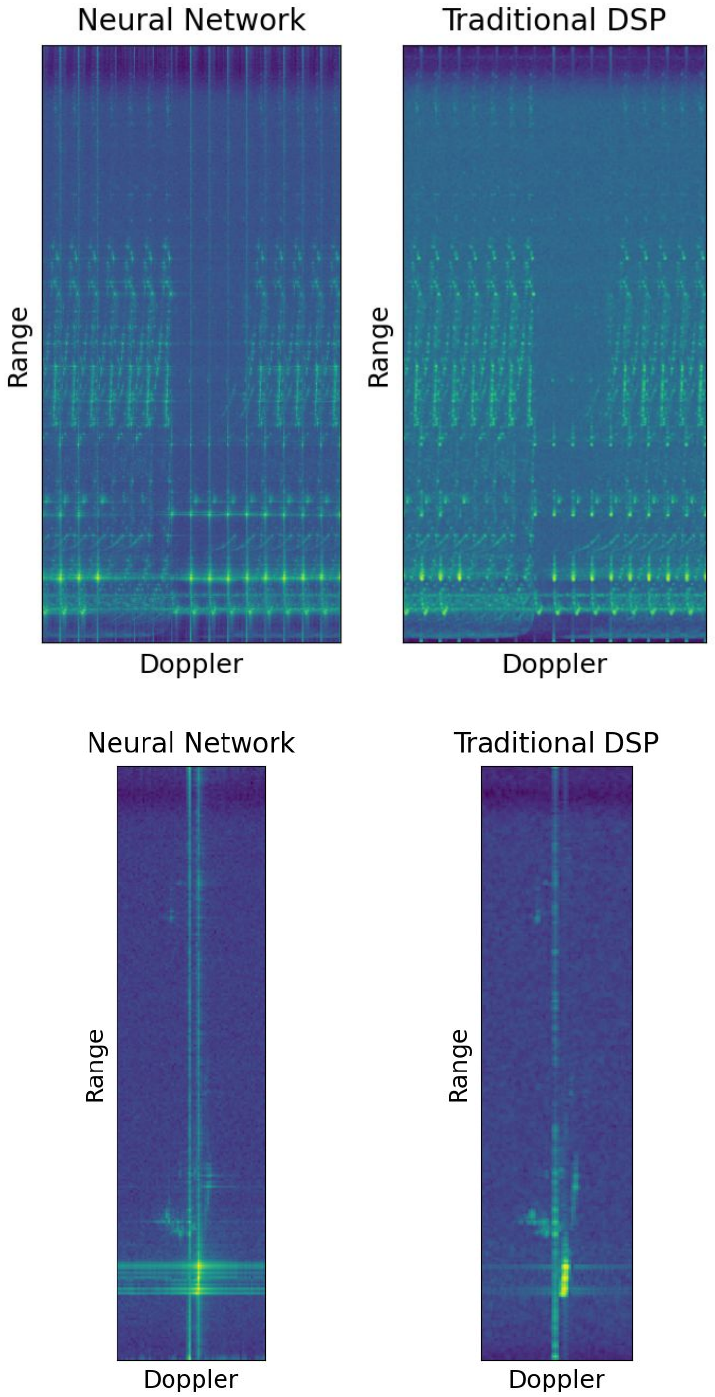}
    \caption{\textbf{Transformation Comparison:} Learned transformations (left column) and traditional RD transformations utilizing Hamming Windows, FFTs, and centering of zero Doppler frequency (right column). HD radar (top row) and LD radar (bottom row).}
    \label{fig:FFT_vs_NN}
\end{figure}

Utilizing linear layers to mimic FFT response is an efficient method of decreasing pre-processing costs, while retaining performance on par or potentially better, depending on the task, than traditional DSP pre-processing and normalization. The layers are able to better encapsulate the radar sensor characteristics in this case. This is encapsulated within the LD radar task of object classification, in which we see utilization of raw ADC and the linear layers provides an increased mAP. In HD radar, utilizing traditional RD inputs performs best, although there is not a significant performance decrease when ADC is utilized. In either case, we note that the transformation learned is no longer Fourier. Fig.\ref{fig:FFT_vs_NN} shows traditional RD FFTs (right column) along with the learned transformation (left column) for both HD (top row) and LD (bottom row) radar.

Being that the Fourier layers can be optimally utilized on GPU, the inference time for a single event decreases in comparison to utilizing RD inputs. If one takes into account the necessary time needed to perform standard FFTs and normalization for RD inputs on HD radar, the average time per one inference is $\sim 156 ms$, whereas with raw ADC the value is one-fifth of this at $\sim 30ms$. LD radar is similar, with the average time per one inference at $\sim 152 ms$ with RAD cubes, $\sim 33 ms$ with RD inputs, and $\sim 20 ms$ with raw ADC inputs. Thus, the utilization of these layers allows the transition toward real-time applications. Note the above values were calculated utilizing an Nvidia RTX 3090 24GB card and Intel~\textregistered ~i7 12700K CPU in which FFTs are allocated to the CPU.

Hamming Windows have shown increased performance in both LD and HD radar datasets due to their ability to reduce side lobe profiles. Centering the zero Doppler frequency, via an \textit{fftshift}, produces performance increases for LD radar, but performance remains approximately the same for HD radar. Due to higher resolution, leakage from positive to negative frequencies does not pose the same impacts as seen in LD radar, where the centering increases performance significantly.

\section{Conclusions}

Radar possesses many desirable traits in relation to autonomous vehicles and other robotic applications, specifically those that rely on the localization of objects in space. The capability to operate in adverse conditions reliably, unlike other emission-based sensors such as LiDAR, make radar a must for fully autonomous systems. Traditional methods of utilizing radar point clouds fall short due to relative sparseness in the generated clouds. To overcome this, methods coupling traditional DSP methods combined with deep learning have been proposed. This too poses new problems, such as the relative computational costs associated with pre-processing inputs utilizing FFTs along with the network growth as input complexity scales. In this work we have shown that we can operate on par, or better depending on the task, utilizing raw ADC inputs via the utilization of fully connected layers that mimic traditional FFTs, yet utilize the efficient and fast computational power of GPUs. These layers require no pre-processing of inputs (including normalization) and can potentially allow the model to learn the physical characteristics of the sensor. This consideration seems to be less prominent in HD radar, where resolution is not an issue. The utilization of dense transformation layers greatly decreases the time associated with a single inference given that no normalization or FFTs are required.

We show the novel application of vision transformers, namely hierarchical Swin Transformers \cite{Liu_2021} to radar-based object detection, utilizing various inputs. Swin transformers utilize W-MHSA which allows the model to retain linear complexity with respect to the input, a strong consideration when deployment on relatively low-power hardware is considered. It is also known that vision transformers scale with training data and as such our contribution is timely given the growing interest in the generation of radar object detection datasets. Our architecture builds off the prior work of Rebut et al.\cite{Rebut_2021}, and is capable of obtaining SOTA results across various datasets with differing sensor configurations, \textit{i.e.} LD and HD radar. The model primarily takes ADC or RD matrices as input and approximates the angular transformation required for object detection, but we have also shown the capability to operate on full RAD cubes with little network alterations, mainly due to fixed-sized embeddings utilized by transformers. This also promotes minimal network growth.


{\small
\bibliographystyle{ieee_fullname}
\bibliography{main}
}

\end{document}